\documentclass[sigconf]{acmart}

\usepackage{booktabs} 
\usepackage{graphicx} 
\usepackage{float} 
\usepackage{subfigure} 
\usepackage{multirow}
\usepackage{multicol}
\usepackage{algorithm}  
\usepackage{algpseudocode} 

\setcopyright{rightsretained}

\acmDOI{10.475/123_4}

\acmISBN{123-4567-24-567/08/06}

\acmConference[ICMR19]{ACM International Conference on Multimedia Retrieval}{June 2019}{Ottawa, Canada}
\acmYear{2019}
\copyrightyear{2019}

\acmArticle{4}
\acmPrice{15.00}

\editor{Jennifer B. Sartor}
\editor{Theo D'Hondt}
\editor{Wolfgang De Meuter}

\begin{document}
\title{Deep Policy Hashing Network with Listwise Supervision}

\author{Shaoying Wang}
\affiliation{
 \institution{School of Electronics and Information Engineering.Sun Yat-Sen University}
  \city{Guangzhou}
 \state{China}
 \postcode{510006}
}
\email{wangshy47@mail2.sysu.edu.cn}

\author{Haijiang Lai}
\affiliation{%
 \institution{Guangdong Key Laboratory of Big Data Analysis and Processing}
 \city{Guangzhou}
 \state{China}
 \postcode{510006}
}
\email{laihanj3@mail.sysu.edu}

\author{Yifan Yang}
\affiliation{%
 \institution{School of Data and Computer Science.Sun Yat-Sen University}
 \city{Guangzhou}
 \state{China}
 \postcode{510006}
}
\email{yangyf26@mail2.sysu.edu.cn}

\author{Jian Yin}
\affiliation{%
 \institution{Guangdong Key Laboratory of Big Data Analysis and Processing}
 \city{Guangzhou}
 \state{China}
 \postcode{510006}
}
\email{issjyin@mail.sysu.edu.cn}

\renewcommand{\shortauthors}{S. Wang et al.}

\begin{abstract}
Deep-networks-based hashing has become a leading approach for large-scale image retrieval, which learns a similarity-preserving network to map similar images to nearby hash codes. The pairwise and triplet losses are two widely used similarity preserving manners for deep hashing. These manners ignore the fact that hashing is a prediction task on the list of binary codes. However, learning deep hashing with listwise supervision is challenging in 1) how to obtain the rank list of whole training set when the batch size of the deep network is always small and 2) how to utilize the listwise supervision. In this paper, we present a novel deep policy hashing architecture with two systems are learned in parallel: a \textit{query network} and a shared and slowly changing \textit{database network}. The following three steps are repeated until convergence: 1) the database network encodes all training samples into binary codes to obtain whole rank list, 2) the query network is trained based on policy learning to maximize a reward that indicates the performance of the whole ranking list of binary codes, e.g., mean average precision (MAP), and 3) the database network is updated as the query network. Extensive evaluations on several benchmark datasets show that the proposed method brings substantial improvements over state-of-the-art hashing methods.
\end{abstract}

%
%
\ccsdesc[500]{Information systems~Image search}

\keywords{Image Retrieval, Hashing, Listwise, Reinforcement Learning, Policy Network}

\maketitle

\section{Introduction}
In the big data era, the amount of images has increased rapidly in social networks and search engines. To guarantee high quality and computation efficiency, hashing methods~\cite{wang2017survey,wang2016learning}, which map high-dimensional media data to compact binary codes such that similar images are mapped to similar binary hash codes, have received considerable attention because of retrieval efficiency.

\begin{figure}[t]
    \centering
    \includegraphics[width=0.45\textwidth]{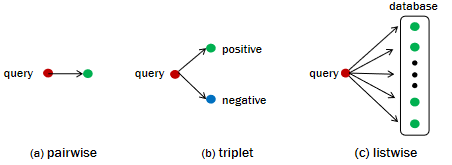}
    \caption{Illustration of three similarity preserving manners: (a) pairwise approaches take image pairs as instances in learning; (b) triplet approaches protect relative similarities among three images, in which the query image is more similar to the positive image than the negative one; (c) our listwise approach takes the query to retrieval the whole data lists in learning.}
    \label{fig:example}
\end{figure}

Many hashing methods have been proposed in the literature. The shallow architectures are firstly proposed to learn the codes. For example, Locality Sensitive Hashing (LSH)~\cite{gionis1999similarity}, Iterative Quantization (ITQ)~\cite{gong2013iterative}, Fast Supervised Discrete Hashing (FSDH)~\cite{gui2018fast}, Spectral hashing (SH)~\cite{weiss2009spectral} use the hand-crafted features. Recently, the deep hashing methods~\cite{cao2017hashnet,zhang2018attention} have achieved impressive results in image retrieval due to the powerful features extracted from the deep networks. Pairwise and triplet losses are two widely used manners to learn the similarity preserving hash functions. The pairwise methods take two images as input and characterize the relationship between the two images, i.e., if the two images are similar, the Hamming distance between the learned binary codes should be small; otherwise, the Hamming distance should be large. The representative works include Deep Asymmetric Pairwise Hashing (DAPH)~\cite{Shen:2017:DAP:3123266.3123345}, Deep Supervised Hashing (DSH)~\cite{liu2016deep}, Deep Supervised Hashing with Pairwise Bit Loss~\cite{Wang:2017:DSH:3094243.3094257} and so on. The triplet approaches~\cite{lai2015simultaneous,zhuang2016fast} preserve relative similarity relations of three images. For instance, given three images $I, I^+, I^-$ that $I$ is more similar to $I^+$ than $I^-$, the goal of triplet loss is to preserve the similarities of the learned binary codes of these three images. 

\begin{figure*}[t]
    \includegraphics[width=1\textwidth]{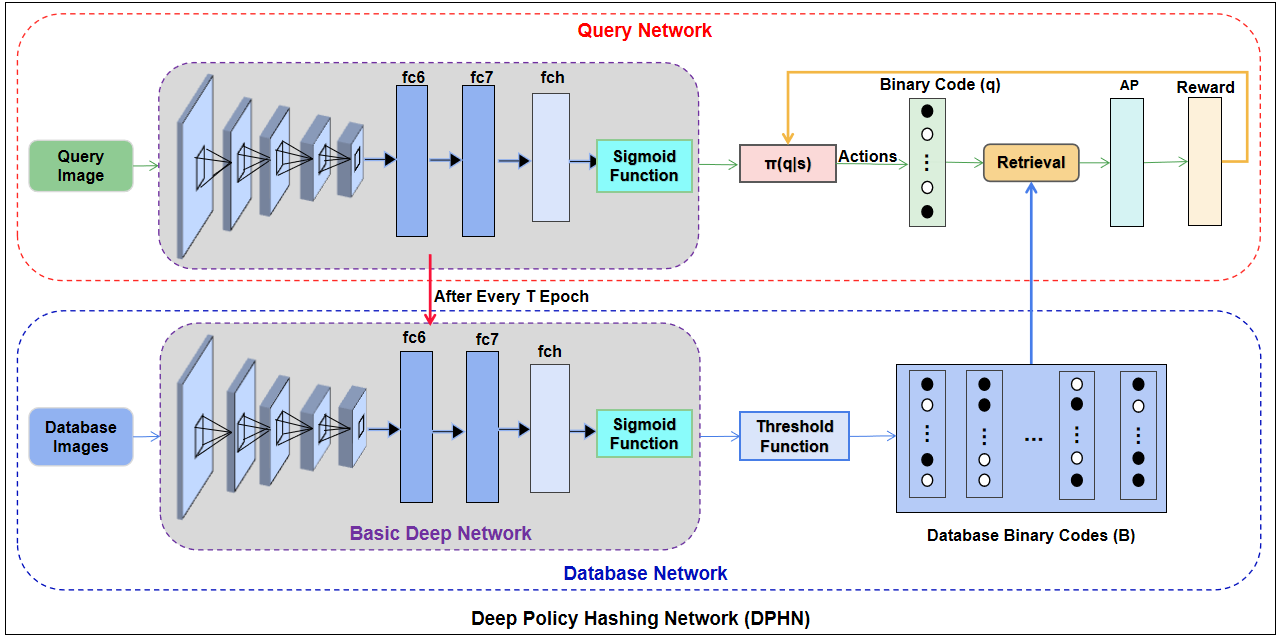}
    \caption{Overview the structure of our propose deep policy hashing network, which consists of two sub-network sharing weights, a query network and a fixed database network. The database network encodes all images in the training set to binary codes, which regarded as the database for retrieval, and the query network maps the query images into binary codes and searches in the database to obtain the average precision (AP) and rewards. Then the gradients of rewards are backpropagated to update the query network and adjust the policy constantly. The weights of the query network are assigned to the database network every $\mathbf{T}$ training epochs to update the database.}
    \label{fig.main}
\end{figure*}

Although the pairwise and triplet approaches offer several advantages, e.g., easy training, these two manners do not consider the fact that the sample lists instead of pairs or triplets are returned when an image is taken as query to retrieve the relevant data. It may result in sub-optimal solution when do not consider the whole rank lists. For example, suppose that $\mathbf{q}$ is a query code and $\{\mathbf{h}_1,\mathbf{h}_2,\mathbf{h}_3,\mathbf{h}_4,\mathbf{h}_5\}$ are the database codes, the query $\mathbf{q}$ and $\{\mathbf{h}_1, \mathbf{h}_2 \}$ are similar and dissimilar with other three codes. Taken $\mathbf{q}$ as the query, two rank lists are returned by two ranked models: $\pi_1 = \{\mathbf{h}_1,\mathbf{h}_3,\mathbf{h}_4,\mathbf{h}_5,\mathbf{h}_2\}$ and $\pi_2 = \{\mathbf{h}_3, \mathbf{h}_1, \mathbf{h}_2, \mathbf{h}_4, \mathbf{h}_5 \}$. Which model performs better? From the pairwise perspective, $\pi_1$ is worse than $\pi_2$ since there are 3 wrong pairs (i.e., $(\mathbf{h}_3,\mathbf{h}_2),(\mathbf{h}_4,\mathbf{h}_2),(\mathbf{h}_5,\mathbf{h}_2)$) in $\pi_1$ and only 2 wrong pairs (i.e., $(\mathbf{h}_3,\mathbf{h}_2),(\mathbf{h}_3,\mathbf{h}_1)$) in $\pi_2$. While when the precision$@1$, i.e., only consider the first returned image, is used as the evaluation metric, $\pi_1$ is better than $\pi_2$. In such case, the pairwise loss does not provide the optimal rank list. 

In this paper, we propose to employ the listwise supervision, in which the whole binary codes are used in learning as illustrated in Figure \ref{fig:example}. Since the deep networks always have a large number of parameters, it is impossible to load all training data into one batch. Hence, the significant questions are then 1) how to obtain the whole binary codes of the training data and 2) how to define a listwise loss function to use the entire binary codes. 

We propose a Deep Policy Hashing Network (DPHN) to address problems. To solve the first problem, we asynchronously execute two hashing networks in parallel inspired by the asynchronous reinforcement learning, e.g., asynchronous advantage actor-critic (A3C)~\cite{mnih2016asynchronous}. These networks have the same network architecture that maps an input image to an approximate hash code, which are referred to \textit{query network} and \textit{database network}. The database network is the copy of the query network. It is similar to asynchronous Q-learning that the parameters in the query network are updated in each mini-batch while the database network is updated later.  In training, the database network is to encode all training data into binary codes. The whole binary codes generated from the database network are used as the retrieval database. Second, to learn the query network with all training codes as the listwise supervision, we adopt the policy gradients to update the query network by performing approximate gradient ascents directly. Especially, for each bit, it has two binary values: 0 or 1. Hence, the query network can be viewed as a policy network that decides which action we should do (move to `0' or `1') for each bit. This policy network can be trained using reinforcement learning to maximize a reward that measures search accuracy when taking the query to retrieve the whole training codes. Since Mean Average Precision (MAP) is the widely used evaluation measure, we adopt it as the reward of the deep policy hashing network. Please note that other evaluation measures can be used as reward. We conduct extensive experiments on three widely used datasets, demonstrating that our proposed approach yields better performance compared with other state-of-the-art methods.

The main contributions of our proposed method can be summarized as follows:
\begin{enumerate}
\item \textbf{The whole ranking list information} is utilized during training the policy network. The whole train data is regarded as a retrieval database during the training process, which ensures that we consider the whole rank list instead of only image pairs or triplets used as instances in learning. 
\item \textbf{A deep policy hashing network} is proposed to learn binary hash codes with listwise supervision directly. The hashing network is viewed as a policy network to generate binary hash codes. Also, we use the evaluation measure, e.g., MAP, as the reward scheme. It will encourage the deep policy hashing network to improve the evaluation measure directly.
\end{enumerate}



\section{RELATED WORKS} 
Existing image hashing methods can be divided into two categories: unsupervised hashing and supervised hashing. The unsupervised methods learn hash network that maps images to binary codes without labels information. Locality Sensitive Hashing (LSH)~\cite{gionis1999similarity} is a typical unsupervised method, which generates binary codes by random linear projection. 
Spectral Hashing (SH)~\cite{weiss2009spectral} uses the relationship between pairwise images to preserve similarity. Iterative Quantization (ITQ)~\cite{gong2013iterative} is proposed to learn hashing function by reducing the quantization gap between real feature space and binary Hamming space. Topology Preserving Hashing (TPH)~\cite{Zhang:2013:TPH:2502081.2502091} tries to preserve the consistent neighbourhood rankings of data by learning a Hamming space. 

Supervised hashing methods are proposed to take advantage of labels information. Predictable Discriminative Binary Code (DBC)~\cite{rastegari2012attribute} learns hash function in hyperplanes and Minimal Loss Hashing (MLH) \cite{norouzi2011minimal} learns hash network by optimizing a hinge-like loss. To deal with linearly inseparable data, Supervised Hashing with Kernels (KSH) \cite{liu2012supervised} and Binary Reconstructive Embbeding (BRE) \cite{kulis2009learning} are proposed. Supervised Discrete Hashing (SDH) \cite{shen2015supervised} improves retrieval accuracy by integrating classify and binary codes generation during training. 
Recently, many deep hashing methods are proposed~\cite{zhu2016deep, yang2018supervised, li2015feature, zhang2018attention, zhuang2016fast, gui2018fast, erin2015deep}. According to the forms of similarity preserving manners, two supervised information are widely used: 1) the pairwise-based methods and 2) the triplet-based methods. The pairwise-based methods take the image pairs as input. For example, Wang et al.~\cite{Wang:2017:DSH:3094243.3094257} proposed a deep supervised hashing network with pairwise labels. The triplet methods consider the relationships among three images. Network in Network Hashing (NINH) \cite{lai2015simultaneous} learns hash network by optimizing triplet loss. Deep hashing methods improve retrieval quality to some extent.
Nevertheless, there still exist some shortcomings which limit higher retrieval quality. These methods only consider the relationship between pairwise or triple images. Our proposed approach will overcome the deficiency by taking whole ranking list information in training.

Very recently, Zhang et.al~\cite{zhang2018deep} proposed a deep reinforcement learning for image hashing, which learns the correlation between different hashing functions. The main difference is that they seek to learn each hashing functions sequentially while our method aims at learning whole ranking list information. Listwise methods are also proposed to learn hash codes. In~\cite{wang2013learning}, the listwise supervision is represented by a set of triplets. Different from that, we directly consider the whole binary codes. 

\section{Deep Policy Hashing Network}

Given an image $I$, our goal is to learn a hashing function $\mathcal{F}:I\rightarrow\{0,1\}^K$, which encodes image $I$ to a $K$-bits compact binary code $\mathcal{F}(I)$. To preserve semantic similarity, the hashing function $\mathcal{F}(\cdot)$ should map similar images to similar compact binary codes in Hamming space, and vice versa. In this paper, we propose a deep policy hashing network as shown in Figure~\ref{fig.main}. The proposed method consists of two ``parallel'' networks: a query network and a database network. These networks have the same architecture that maps an input image to a hash code. The database network encodes all training images into binary codes, which is used to obtain the whole list of codes. To utilize the listwise supervision, the query network is learned with all training codes via reinforcement learning. We will present the detail of these two networks in the following parts.

\subsection{Database Network}
As shown in Figure~\ref{fig.main}, the database network consists of a basic deep network and a threshold function. The deep network includes stacked convolutional and fully-connected layers, followed by a sigmoid function that generates the intermediate features. To extract powerful feature representations, we adopt the VGG-19 network \cite{simonyan2014very} as the basic architecture. The first 18 layers follow the same settings in the VGG-19 network, while the last fully-connected layer (fch) is replaced by a $K$-dimensional fully connective layer, where $K$ is the bits of binary codes. Then, a sigmoid function is added to restrict the values in the range $[0,1]$. We denote the output of the basic deep network as $\mathbf{s} = \mathcal{F}_D(I)$, where $I$ is an input image. 
 
Further, we also add a threshold function to generate binary codes of image $I$ as
\begin{equation}
    b_k =  \left\{
    \begin{array}{rcl}
    1	&	&{s_k \geq 0.5}\\
    0	&	&otherwise
\end{array}\right.,
\end{equation}
where $s_k$ is the $k$-th entry of $\mathbf{s}$, and $b_k$ is the $k$-th binary code of the image $I$.

Given $n$ training samples $\{I^{(i)} \}_{i=1}^n$, we can obtain $n$ binary codes $\mathbf{B} = \{\mathbf{b}^{(i)}\}_{i=1}^n$. These codes are used as a retrieval database when training the query network, which will be described in the Subsection~\ref{query_network}.

\subsection{Query Network} 
\label{query_network}   

Given a query, taking all retrieval samples into account has not been well-explored in the most existing deep hashing methods. In this paper, we consider the advantages of reinforcement learning and design a policy network to learn with listwise supervision.

We consider the whole training data into the training phase, which aims to learn the hashing network as shown in Figure~\ref{fig.main}. The query network considers two inputs: 1) the query image and 2) the retrieval database binary codes $\mathbf{B} = \{\mathbf{b}^{(i)}\}_{i=1}^n$ (generated by the database network). Formally, for each mini-batch in the training phase, an image $I$ goes through the query network and is encoded as a query code $\mathbf{q}$. Note that the query network uses the same basic deep network as the database network, which is also built on the VGG-19 layer net. The symbol $\mathbf{q} \to \mathbf{B}$ denotes the code $\mathbf{q}$ is taken as the query to retrieve the relevant data in the retrieval database $\mathbf{B}$. Our query network works as follows: we generate query codes to retrieval the database and aim to obtain good performance. A policy learning is proposed to learn the network. The state, actions and reward of our proposed model are shown in detail as follows.

\textbf{Actions:} Since each bit has only two values, i.e., `0' or `1', the number of actions for each bit are two: `0' or `1'. For $K$-bit binary code, there are $2^K$ actions in total. The deep hashing network can be viewed to decide the value to be `0' or `1' for each bit. 

\textbf{State:} The input image is considered as an observing state, which provides enough information for the agent to determine the actions.

\textbf{Reward:} Since average precision (AP)~\footnote{MAP is mean AP for all query images. It is only one query, so, the AP but not MAP is used.} is a widely used evaluation measure. Hence, AP is applied to guide the agent to learn the policy network.

\begin{algorithm}[htp]  
  \caption{The Pseudo Code for Our approach DPHN}  
  \begin{algorithmic}[1]  
    \State Initialize the first 18 layers of Basic Deep Network (BDN) with the weights of the pre-trained VGG-19 network, and initialize the last layer randomly
    \Repeat
        \State Perform a gradient descent step on $L_{triplet}$ w.r.t the wieghts of BDN defined as $\mathbf{W_b}$
    \Until $BDN$ is convergent
    \State Initialize the Query Network (QN) with weights $\mathbf{W_q} = \mathbf{W_b}$
    \State Initialize the Database Network (DN) with weights $\mathbf{W_d} = \mathbf{W_q}$
    \State Regard all training data as database images for retrieval
    \State Generate database binary codes $\mathbf{B}$ with DN
    \For{$epoch = 1, M$}
        \State Generate query binary code $\mathbf{q} \sim Bernoulli(\mathbf{s})$ with QN
        \State $\mathbf{q} \rightarrow \mathbf{B}$
        \State Compute $AP$ and reward $R(\mathbf{q})$
        \State Compute the overall loss $L$ with Equation \ref{eqn:loss}
        \State Update QN with gradients computed with Equation \ref{eqn:gradients}
        \If{$epoch~\% ~T == 0$}
            \State reset $\mathbf{W_d} = \mathbf{W_q}$
            \State update $\mathbf{B}$ with new DN
        \EndIf
    \EndFor
  \end{algorithmic}  
  \label{alg:DPHN}
\end{algorithm} 

\subsection{Optimization}

In training, the database network encodes all training data into binary codes $\mathbf{B}$, after which the query network is learnt via policy gradients. The database network is updated as the query network every $\mathbf{T}$ training epochs. Thus, we only need to optimize the parameters in the query network. In this section, we will show how to train the query network. 

Unlike standard reinforcement learning algorithms, we train the policy to predict all actions at once, which can be viewed as a single-step Markov Decision Process (MDP)~\cite{sutton2011reinforcement}. Given an image $I$, inspired by the idea in BlockDrop \cite{wu2018blockdrop}, we define the policy of binarized behaviour as a $K$-dimensional Bernoulli Distribution:
\begin{equation}
    \pi_\mathbf{W}(\mathbf{q} |s) = \prod_{k=1}^{K}s_k^{q_k}(1-s_k)^{1-q_k},
\end{equation}
where $K$ is the bits of binary codes, $\mathbf{W}$ is the parameters of the query network, $s_k \in [0,1]$ is the $k$-th value of the intermediate feature, and $q_k \in \{0,1\}$ is the action for the $k$-th bit, which is also the binary code for image $I$. 

Given a query image $I$, the reward function is:
\begin{equation}
    R(\mathbf{q}) = \left\{
    \begin{array}{rcl}
    AP(\mathbf{q})	&	&{AP(\mathbf{q}) > \beta}\\
    AP(\mathbf{q}) - 1	&	&otherwise
\end{array}\right.,
\label{eqn:reward}
\end{equation}
where $\mathbf{q}$ is the binary code of the image $I$, and $AP$ is evaluation value for $\mathbf{q} \to \mathbf{B}$ which takes $\mathbf{q}$ as a query to retrieval the database $\mathbf{B}$, which is defined as
\begin{equation}
    AP(\mathbf{q}) = \frac{1}{N^+}\sum_{j=1}^{n}\frac{N^+_j}{j}\cdot sim(\mathbf{q},\mathbf{b}^{(j)}),
\end{equation}
where $n$ is the number of images in the database, $N^+$ is the total number of similar images in database $\mathbf{B}$ w.r.t the query $\mathbf{q}$. And $N^+_j$ is the number of similar images in the top $j$ images, and $sim(\mathbf{q},\mathbf{b}^{(j)})=1$ if the $j$-th returned image is similar to the query, and otherwise $sim(\mathbf{q},\mathbf{b}^{(j)})=0$. Note that the data in database $\mathbf{B}$ are binary codes, it is efficient to calculate the AP. The $\beta$ determines whether the agent should be rewarded or punished. If $AP > \beta$, the agent will receive a positive reward. Otherwise, it will receive a negative reward. The smaller the AP, the greater the penalty. 
Our goal is to maximize the following expected reward:
\begin{equation}
    J = \mathbb{E}_{\mathbf{q}\sim\pi_\mathbf{W}}[R(\mathbf{q})].
\end{equation}
By utilizing policy gradient  \cite{williams1992simple}, we can calculate the gradients of the expected reward and update the parameters of the query network with backpropagation algorithm. We define the gradients as the following equation:
\begin{equation}
\begin{split}
    \bigtriangledown_{\mathbf{W}}J &= \mathbb{E}[R(\mathbf{q})\bigtriangledown_{\mathbf{W}}log\prod_{k=1}^K s_k^{q_k}(1-s_k)^{1-q_k}]\\
    &= \mathbb{E}[R(\mathbf{q})\bigtriangledown_{\mathbf{W}}\sum_{k=1}^K log(s_k^{q_k} + (1-s_k)^{1-q_k})].
\end{split}
\end{equation}
We adopt Monte-Carlo algorithm to sample data and replace the expected gradients by estimated gradients. The estimated gradients are unbiased but with high variance. A self-critical baseline $R(\bar{\mathbf{q}})$ is utilized to reduce variance, and the gradients can be modified as:
\begin{equation}
\begin{split}
    \bigtriangledown_{\mathbf{W}}J &\approx \mathbb{E}[D\bigtriangledown_{\mathbf{W}}\sum_{k=1}^K log(s_k^{q_k} + (1-s_k)^{1-q_k})]\\
    &\approx \frac{1}{N}\sum_{i=0}^N D\bigtriangledown_{\mathbf{W}}\sum_{k=1}^K log(s_k^{b_k} + (1-s_k)^{1-b_k}),
\end{split}
\end{equation}
where $N$ is the batch size, $D=R(\mathbf{q})-R(\bar{\mathbf{q}})$ and $\bar{\mathbf{q}}$ is selected based on $\mathbf{s}$ with a threshold function: $\bar{\mathbf{q}}=0$ if $\mathbf{s} < 0.5$ and $\bar{\mathbf{q}}=1$ otherwise. For the unity of expression, we defined the overall policy loss as:
\begin{equation}
    L_{policy} = -J.
\end{equation}


However, it is difficult to guarantee the convergence of the policy network. We joint the policy loss with the triplet ranking loss to accelerate the convergence of network. The overall loss function can be written as:
\begin{equation}
    L = L_{triplet} + L_{policy},
\label{eqn:loss}
\end{equation}
where the triplet ranking loss is defined as:
\begin{equation}
\begin{split}
    L_{triplet}(\mathbf{s},\mathbf{s}^+,\mathbf{s}^-)
    = \max\{0, m + \|\mathbf{s}- \mathbf{s}^+\|_2^2 - \|\mathbf{s}- \mathbf{s}^-\|_2^2\},\\
\end{split}
\label{eqn:triplet}
\end{equation}
where $m > 0$ is a margin hyper parameter and $\mathbf{s}$,$\mathbf{s}^+$,$\mathbf{s}^-$ are the intermediate features of training image $I$,$I^+$,$I^-$, where image $I$ is more similar to $I^+$ than to $I^-$. Then the overall gradients can be calculated with the following equation:
\begin{equation}
\begin{split}
    \bigtriangledown_{\mathbf{W}}L &= \bigtriangledown_{\mathbf{W}}L_{triplet} + \bigtriangledown_{\mathbf{W}}L_{policy}\\
    &= \bigtriangledown_{\mathbf{W}}L_{triplet} - \bigtriangledown_{\mathbf{W}}J,\\
\end{split}
\label{eqn:gradients}
\end{equation}
where the detail calculation of $\bigtriangledown_{\mathbf{W}}L_{triplet}$ has been explained in \cite{lai2015simultaneous}, and we will not repeat here. Algorithm \ref{alg:DPHN} shows the complete training process for our approach DPHN.

\section{Experiments}
In this section, we will present our experiments on three image retrieval datasets and the results compared with several state-of-the-art hashing methods, including DRLIH \cite{zhang2018deep}, HashNet \cite{cao2017hashnet}, NINH \cite{lai2015simultaneous}, CNNH \cite{xia2014supervised}, DSH \cite{liu2016deep}, LSH \cite{gionis1999similarity}, SDH \cite{shen2015supervised} and SH \cite{weiss2009spectral}.

\subsection{Datasets}
The experiments are conducted on three benchmark image retrieval datasets: CIFAR-10 \cite{kulis2009learning}, NUS-WIDE \cite{Chua:2009:NRW:1646396.1646452} and MIRFlickr \cite{Huiskes:2008:MFR:1460096.1460104}.
\begin{itemize}
    \item \textbf{CIFAR-10} consists of $60,000$ color images from $10$ classes with the size of $32\times32$. There are $6,000$ images per class. To fairly compare with other methods, we randomly select $100$ images per class as query set and 500 images per class as training set as suggested by~\cite{zhang2018deep}. All images except those in the query set are selected as the retrieval database.
    \item \textbf{NUS-WIDE} is a public web image dataset which consists of $269,648$ images associated with one or multiple labels from 81 categories. Following~\cite{zhang2018deep}, we filter $21$ most common categories which each category has $5,000$ images at least. The $100$ images per category are selected as the query set (totally $2,100$ images), and $500$ images per category are selected as the training set (totally $10,500$ images). All images except the query set are regarded as the database.
    \item \textbf{MIRFlickr} is a dataset which contains $25,000$ images selected from Flickr, and each image is labelled from $38$ categories. Following the setting of~ \cite{zhang2018deep}, $1,000$ images and $5,000$ images are selected as the query set and the training set, respectively. Apart from the images in the query set, the rest images are considered as the database set.
\end{itemize}

\begin{table*}[t]
    \caption{MAP scores with different length of binary codes on CIFAR-10, NUS-WIDE and MIRFlickr datasets. The MAP scores of NUS-WIDE dataset are calculated based on top 5000 returned images, while the MAP scores of another two datasets are based on all returned images. The best results are shown in bold.}
    \label{tab:MAP}
    \begin{tabular}{c|cccc|cccc|cccc}
    \hline
    \multirow{2}{*}{\textbf{Methods}} & \multicolumn{4}{c|}{\textbf{CIFAR10}}   & \multicolumn{4}{c|}{\textbf{NUS-WIDE}}  & \multicolumn{4}{c}{\textbf{MIRFlickr}} \\ 
    \cline{2-13} & \textbf{12bits} & \textbf{24bits} & \textbf{32bits} & \textbf{48bits} & \textbf{12bits} & \textbf{24bits} & \textbf{32bits} & \textbf{48bits} & \textbf{12bits} & \textbf{24bits} & \textbf{32bits} & \textbf{48bits} \\ \hline
    \textbf{DPHN (ours)} & \textbf{0.844}  & \textbf{0.862}  & \textbf{0.868}  
    & \textbf{0.878}  & \textbf{0.834}  & \textbf0.849   & \textbf{0.850}  & \textbf{0.854} 
    & \textbf{0.827}  & \textbf{0.840}  & \textbf{0.837}  & \textbf{0.847}  \\
    \textbf{HashNet}& 0.765  & 0.823  & 0.840  & 0.843  & 0.812  & 0.833  & 0.830   & 0.840  & 0.777  & 0.782  & 0.785  & 0.785  \\
    \textbf{DSH}    & 0.708  & 0.712  & 0.751  & 0.720  & 0.793  & 0.804  & 0.815   & 0.800  & 0.651  & 0.681  & 0.684  & 0.686  \\
    \textbf{NINH}   & 0.792  & 0.818  & 0.832  & 0.830  & 0.808  & 0.827  & 0.827   & 0.827  & 0.772  & 0.756  & 0.760  & 0.778  \\
    \textbf{CNNH}   & 0.683  & 0.692  & 0.667  & 0.623  & 0.768  & 0.784  & 0.790   & 0.740  & 0.763  & 0.757  & 0.758  & 0.755  \\ \hline
    \textbf{SDH-VGG19}& 0.430  & 0.652  & 0.653  & 0.665  & 0.730  & 0.797  & 0.819  & 0.830  & 0.762  & 0.739  & 0.737  & 0.747 \\
    \textbf{SH-VGG19} & 0.224  & 0.213  & 0.213  & 0.209  & 0.712  & 0.697  & 0.689  & 0.682  & 0.618  & 0.604  & 0.598  & 0.595  \\
    \textbf{LSH-VGG19}& 0.133  & 0.171  & 0.178  & 0.198  & 0.518  & 0.567  & 0.618  & 0.651  & 0.575  & 0.584  & 0.604  & 0.614  \\ \hline
    \textbf{SDH}& 0.255  & 0.330  & 0.344  & 0.360  & 0.460  & 0.510  & 0.519  & 0.525  & 0.595  & 0.601  & 0.608  & 0.605  \\
    \textbf{SH} & 0.124  & 0.125  & 0.125  & 0.126  & 0.452  & 0.445  & 0.443  & 0.437  & 0.561  & 0.562  & 0.563  & 0.562   \\
    \textbf{LSH}& 0.116  & 0.121  & 0.124  & 0.131  & 0.436  & 0.414  & 0.432  & 0.442  & 0.557  & 0.564  & 0.562  & 0.569   \\ \hline
    \end{tabular}
\end{table*}

\subsection{Experiment Settings and Evaluation Metrics }
The implementation of our proposed method is based on the deep learning framework: PyTorch~\footnote{https://pytorch.org/}. We utilize stochastic gradient descend with $0.9$ momentum as the optimizer. The parameters of the first $18$ layers are initialized with the pre-trained VGG-19 model on ImageNet dataset \cite{russakovsky2015imagenet}. The inputs of our network are $224\times224$ raw images and their corresponding labels. We set the initial learning rate to be $0.001$ and decrease it by a factor of $10$ every $50$ epochs. The weight decay are fixed to be $0.0005$. The $\beta$ in reward function is set to be $0.4$, which controls the balance between the reward and punishment. For the margin $m$ in the triplet ranking loss, we set it to $1, 2, 2, 4$ when the binary bits is $12, 24, 32$ and $48$ respectively. The batch size is $50$, and the database network is updated every $50$ epoch.

We compare our method with $8$ state-of-the-art hashing methods, including the deep supervised hashing methods and the unsupervised methods. For deep hashing methods, raw images are used as input. For a fair comparison, the same VGG-19 network architecture is utilized as the basic structure for all deep hashing methods. We also show the results of the shallow hashing methods with two different features: deep features and hand-crafted features. The deep features are the $4096$-dimensional vectors that are extracted from pre-trained VGG-19 models. Following~\cite{zhang2018deep}, the 512-dimensional GIST features are utilized for CIFAR-10 and MIRFlickr and 500-dimensional bag-of-words features for NUS-WIDE. For a fair comparison, the results of all other hashing methods are cited from \cite{zhang2018deep} directly.

We evaluate image retrieval performance based on three standard evaluation metrics: Mean Average Precision ($MAP$), Precision within Hamming distance 2 ($P@H \leq 2$) and Precision at top $K$ ($P@K$). Mean Average Precision ($MAP$) is the mean of average precision ($AP$) computed in Equation~\ref{tab:MAP}. Precision within Hamming distance 2 ($P@H \leq 2$) is the precision of returned images with Hamming distance within 2 using hash search. This standard evaluation metric is significant for retrieval effectiveness since Hamming ranking require only $O(1)$ time for each query. Precision at top $K$ ($P@K$) is the precision concerning the different number of top $K$ returned samples from ranking list.

\begin{figure*}[t]
    \centering
    \begin{subfigure}
    \centering
    \includegraphics[height=1.52in]{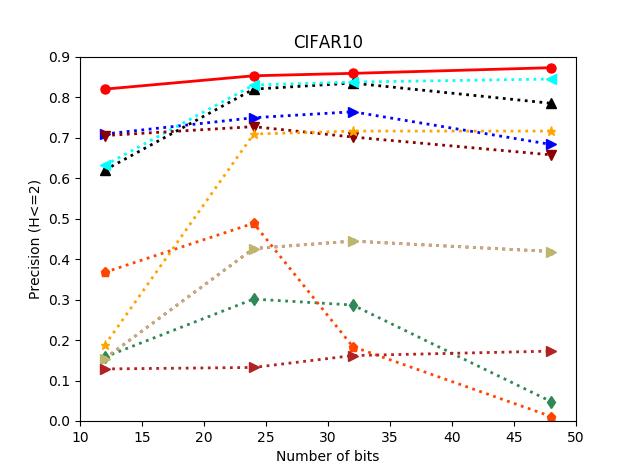}
    \end{subfigure}
    \begin{subfigure}
    \centering
    \includegraphics[height=1.52in]{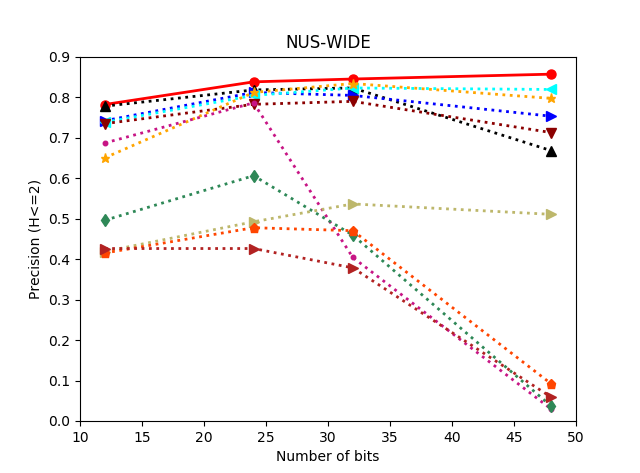}
    \end{subfigure}
    \begin{subfigure}
    \centering
    \includegraphics[height=1.52in]{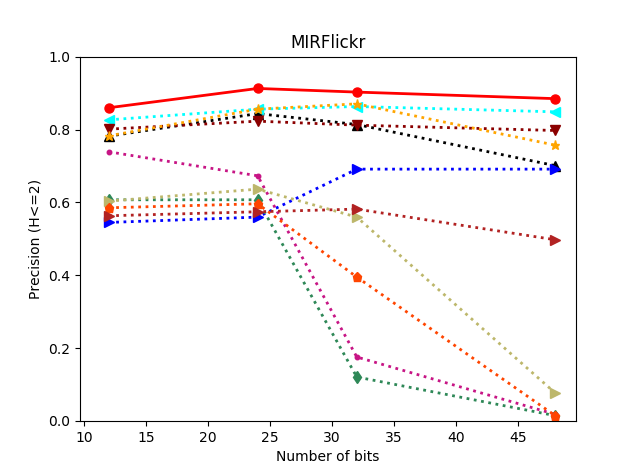}
    \end{subfigure}
    \begin{subfigure}
    \centering
    \includegraphics[height=1.2in]{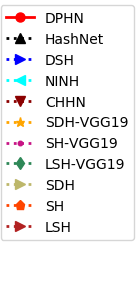}
    \end{subfigure}
    \caption{Precision within Hamming distance 2 using hashing lookup on three datasets.}
    \label{fig:p2}
\end{figure*}

\begin{figure*}[t]
    \centering
    \begin{subfigure}
    \centering
    \includegraphics[height=1.52in]{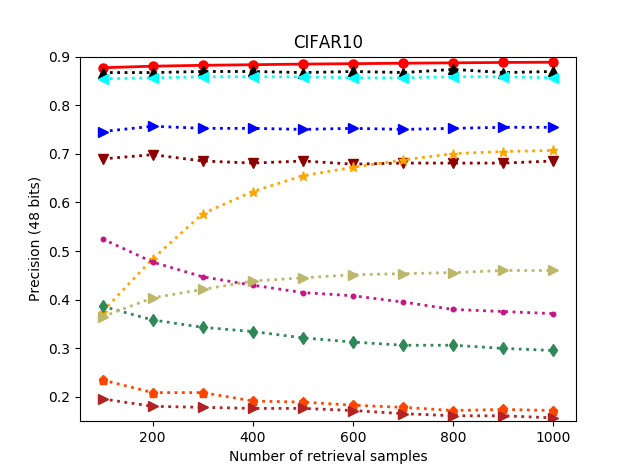}
    \end{subfigure}
    \begin{subfigure}
    \centering
    \includegraphics[height=1.52in]{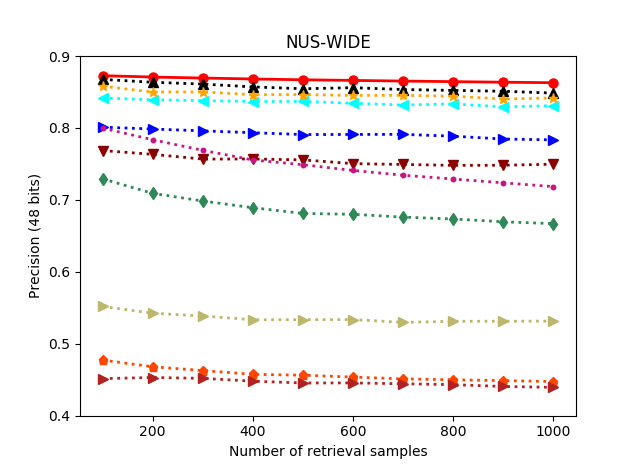}
    \end{subfigure}
    \begin{subfigure}
    \centering
    \includegraphics[height=1.52in]{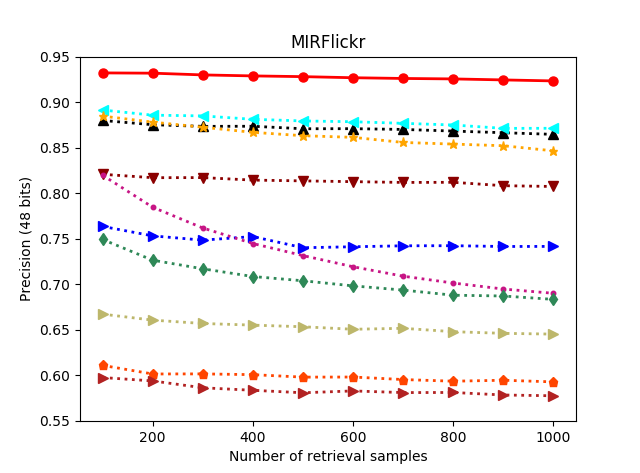}
    \end{subfigure}
    \begin{subfigure}
    \centering
    \includegraphics[height=1.2in]{legends.png}
    \end{subfigure}
    \caption{Precision at top K returned results on three datasets.}
    \label{fig:topk}
\end{figure*}

\subsection{Comparison with State-of-the-art Methods}
In this set of experiments, we evaluate and compare the proposed method with several state-of-the-art methods.

Table~\ref{tab:MAP} shows the comparison results of the MAP on three databases. It is divided into three parts: deep hashing methods, shallow methods with deep features extracted by VGG-19 models and shallow methods with hand-crafted features. All these methods are pairwise, or triplet approaches, except ours. It can be observed that the proposed DPHN performs significantly better than all baselines. Specifically, on CIFAR-10 dataset, DPHN achieves a MAP of 0.844 on 12 bits, which shows an increment of about 8\% over the second best method HashNet. Similar to CIFAR-10, DPHN also performs better than other methods on NUS-WIDE dataset. DPHN achieves a MAP of 0.834 on 12 bits, compared with
0.812 of HashNet. On MIRFlickr dataset, the MAP of our approach DPHN is 0.847 on 48 bits, which shows increments of 6\% over HashNet, 10\% over SDH-VGG19 and 24\% over SDH. 

Figure~\ref{fig:p2} shows the results of precision within Hamming distance 2 ($P@H \leq 2$) on three datasets. We can see that DPHN almost achieves the best performance on three databases. Even with short binary codes, like 12-bits, DPHN achieves best performance than other methods, demonstrating that our approach learns more compact information. The precisions of some methods decrease when using longer binary codes since many queries fail to return images within Hamming radius 2. While our approach achieves higher precision with longer binary codes, the reason may be that our method optimizes the evaluation measure directly.

Figure~\ref{fig:topk} shows the compared results of the precision at top K. Again, DPHN achieves the best performance compared with state-of-the-art methods. Notably, the average precision at top K of DPHN is higher than 0.9 on MIRFlickr dataset, which outperforms other methods. 

\begin{table*}[t]
    \caption{The comparing results of MAP scores between baseline and our proposed approach with different length of binary codes on CIFAR10, NUS-WIDE and MIRFlickr datasets. The best results are shown in bold.}
    \label{tab:baseline}
    \begin{tabular}{c|cccc|cccc|cccc}
    \hline
    \multirow{2}{*}{\textbf{Methods}} & \multicolumn{4}{c|}{\textbf{CIFAR10}}   & \multicolumn{4}{c|}{\textbf{NUS-WIDE}}  & \multicolumn{4}{c}{\textbf{MIRFlickr}} \\ 
    \cline{2-13} & \textbf{12bits} & \textbf{24bits} & \textbf{32bits} & \textbf{48bits} & \textbf{12bits} & \textbf{24bits} & \textbf{32bits} & \textbf{48bits} & \textbf{12bits} & \textbf{24bits} & \textbf{32bits} & \textbf{48bits} \\ \hline
    \textbf{DPHN (ours)} & \textbf{0.844}  & \textbf{0.862}  & \textbf{0.868}  
    & \textbf{0.878}  & \textbf{0.834}  & \textbf{0.849}  & \textbf{0.850}  & \textbf{0.854} 
    & \textbf{0.827}  & \textbf{0.840}  & \textbf{0.837}  & \textbf{0.847}  \\
    \textbf{DRLIH}   & 0.816  & 0.843  & 0.855  & 0.853  & 0.823  & {0.846}  & 0.845  & 0.853  & 0.796  & 0.811  & 0.810  & 0.814\\ \hline
 \end{tabular}
\end{table*}

\begin{figure*}[th]
    \centering
    \begin{subfigure}
    \centering
    \includegraphics[height=1.52in]{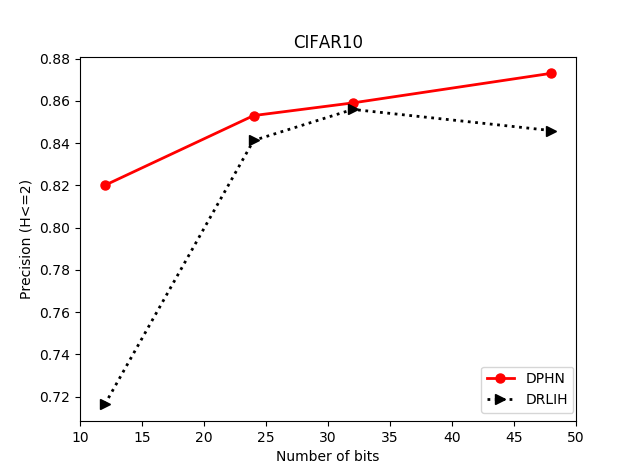}
    \end{subfigure}
    \begin{subfigure}
    \centering
    \includegraphics[height=1.52in]{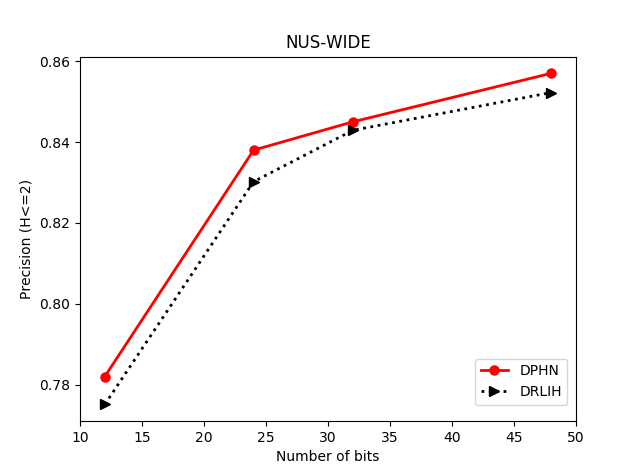}
    \end{subfigure}
    \begin{subfigure}
    \centering
    \includegraphics[height=1.52in]{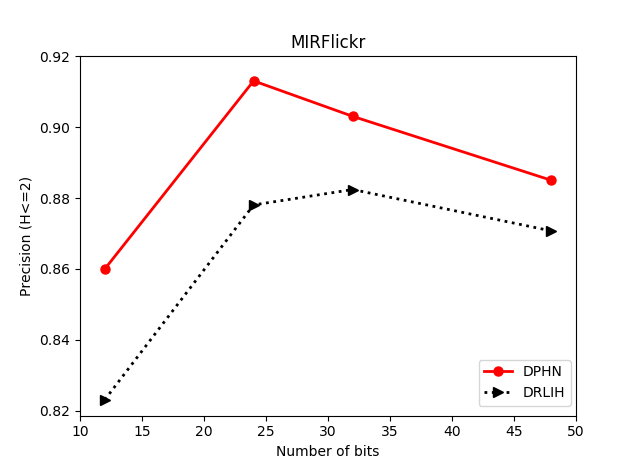}
    \end{subfigure}
    \caption{Precision within Hamming distance 2 using hashing lookup compared with baseline method on three datasets.}
    \label{fig:p2b}
\end{figure*}

\begin{figure*}[th]
    \centering
    \begin{subfigure}
    \centering
    \includegraphics[height=1.52in]{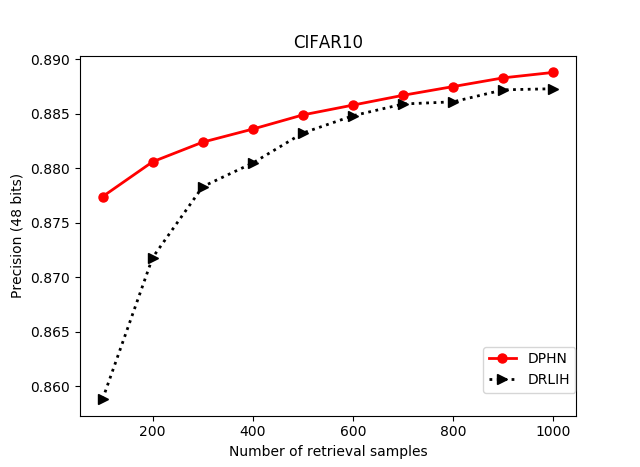}
    \end{subfigure}
    \begin{subfigure}
    \centering
    \includegraphics[height=1.52in]{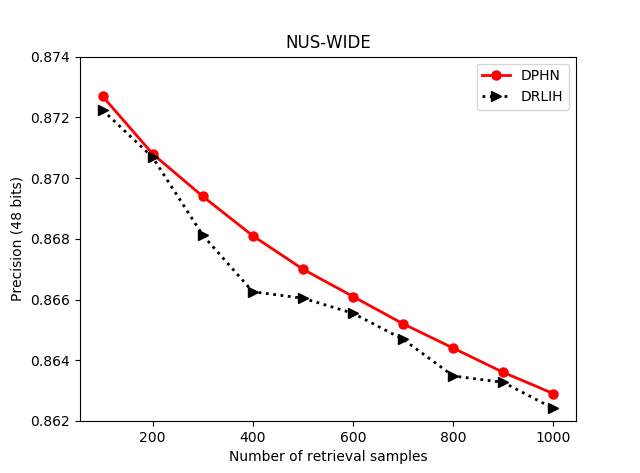}
    \end{subfigure}
    \begin{subfigure}
    \centering
    \includegraphics[height=1.52in]{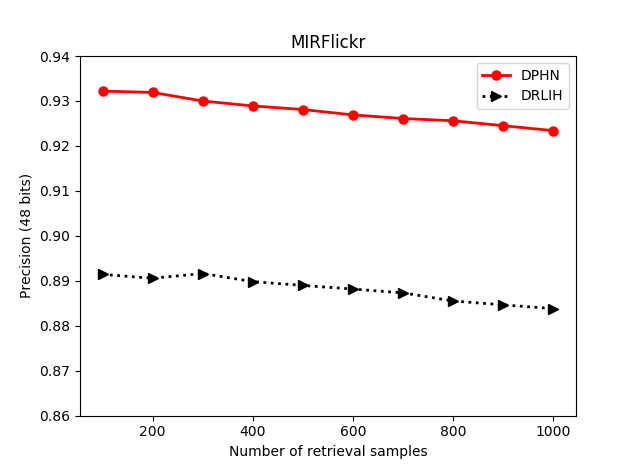}
    \end{subfigure}
    \caption{Precision at top K returned results compared with baseline method on three datasets.}
    \label{fig:topkb}
\end{figure*}

\subsection{Comparison with the Policy Learning Method}
In this set of our experiments, we do ablation study to clarify the impact of the policy learning of our method on the final performance.
DRLIH, a reinforcement learning method with a different policy from ours, is selected as the baseline. The purpose of our strategy is to learn whole ranking list information and improve evaluation measure directly, while DRLIH proposed a policy to capture ranking errors and learn each hashing function sequentially. These comparisons can answer us whether our policy learning can contribute the accuracy or not.

Table~\ref{tab:baseline} shows the comparison results of MAP. On CIFAR10 dataset, the average MAP of DPHN is 0.863, which achieves an improvement of 2\% compared with the average MAP of the baseline (average 0.842). Our proposed approach improves the average MAP from 0.842 to 0.845 and from 0.808 to 0.838 on NUS-WIDE dataset and MIRFlickr, respectively. Therefore, the comparison results prove the effectiveness of our policy learning for improving the quality of image retrieval.

Figure~\ref{fig:p2b} shows the precision within Hamming distance 2 compared with the baseline method. DPHN performs better than DRLIH, especially on MIRFlickr dataset. We can observe the same trend in precision at top K returned samples shown on Figure~\ref{fig:topkb}. The results shown above demonstrate that our policy performs better than the policy of the baseline method.

\begin{figure}[htp]
    \centering
    \includegraphics[width=0.45\textwidth]{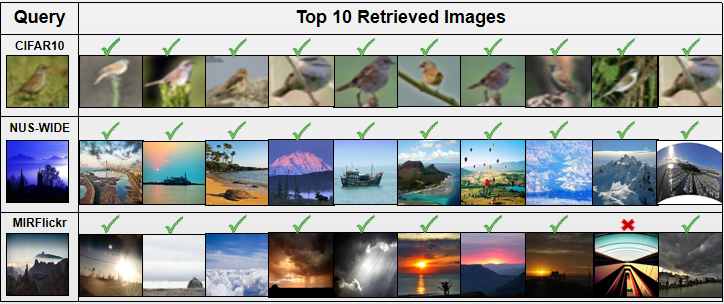}
    \caption{The top 10 images returned by DPHN using Hamming ranking on 48bits binary codes.}
    \label{fig:returnimgs}
\end{figure}

In Figure~\ref{fig:returnimgs}, we visualize the top 10 returned images of three query images on CIFAR10, NUS-WIDE, MIRFlickr with our approach DPHN. It shows that our approach can obtain satisfactory results.

Two observations can be made from comparison results mentioned above: 1) our proposed method with listwise supervision can achieve better performance than the pairwise techniques, e.g., DSH, the triplet methods, e.g., NINH. It is desired to take listwise supervision in training. And 2) our deep policy network also performs better than other reinforcement learning method DRLIH.

\section{CONCLUSION}
In this paper, we proposed DHPN, a hashing method to encode images into binary codes for effective retrieval. First of all, a database network was proposed to encode all images in training set into binary codes. Secondly, a policy network was further proposed to take the whole binary codes into account. Since the reward of the policy network is associated with AP, we could improve MAP directly. Experiments conducted on three widely used datasets proved that our approach achieves higher retrieval quality than existing hashing methods.

\bibliographystyle{ACM-Reference-Format.bst}
\bibliography{references.bib}

\end{document}